\newif\ifanonymized
\anonymizedfalse
\documentclass[10pt]{article}
\ifanonymized
  \usepackage{tmlr}
\else
  \usepackage[preprint]{tmlr}
\fi

\usepackage[utf8]{inputenc}
\usepackage{microtype}
\usepackage{amsmath,amssymb,mathtools}
\usepackage{booktabs,tabularx,multirow}
\usepackage{enumitem}
\usepackage{xcolor}
\usepackage{graphicx}
\usepackage{float}
\usepackage{tikz}
\usetikzlibrary{arrows.meta,fit,positioning}
\usepackage[hidelinks]{hyperref}
\usepackage[nameinlink,noabbrev]{cleveref}
\ifanonymized
  \newcommand{\pdfauthorvalue}{Anonymous Author(s)}
  \author{\name{Anonymous Author(s)}}
\else
  \newcommand{\pdfauthorvalue}{Guodong Xu}
\newcommand{\namedauthorstatement}{%
\textbf{Author and affiliation.} Guodong Xu is affiliated with Qingdao
Guodongxiansheng Network Technology Co., Ltd. (Gu\v{o}d\`ong X\=iansheng).
Correspondence: \texttt{kzkz137806@gmail.com}.%
}
\author{%
\name{Guodong Xu}\\
\addr{Qingdao Guodongxiansheng Network Technology Co., Ltd.}\\
\addr{(Gu\v{o}d\`ong X\=iansheng)}\\
\email{kzkz137806@gmail.com}%
}

\fi
\hypersetup{
  pdftitle={Calibrating Criterion Revision in LLM Agents: Failure Modes and a Trace-Anchored Protocol},
  pdfauthor={\pdfauthorvalue},
  pdfsubject={Criterion revision, persistent memory, intervention-based calibration, and trace-anchored evaluation in LLM agent systems},
  pdfkeywords={LLM agent systems, criterion revision, persistent memory, intervention, scorer validation}
}

\newcolumntype{P}[1]{>{\raggedright\arraybackslash}p{#1}}
\newcolumntype{R}[1]{>{\raggedleft\arraybackslash}p{#1}}
\newcolumntype{Y}{>{\raggedright\arraybackslash}X}
\definecolor{cmblue}{HTML}{2E73B8}
\definecolor{cmfill}{HTML}{EAF3FA}
\definecolor{cmorange}{HTML}{B96C10}
\definecolor{cmorangefill}{HTML}{FFF4E5}
\definecolor{cmred}{HTML}{B52222}
\definecolor{cmgray}{HTML}{536273}
\newcommand{\cmb}{CMB-0.1}
\newcommand{\cmbfour}{CMB-0.4}
\newcommand{\code}[1]{\texttt{#1}}

\title{Calibrating Criterion Revision in LLM Agents:\\
Failure Modes and a Trace-Anchored Protocol}
\date{}

\begin{document}
\maketitle

\begin{abstract}
Language-model agents can improve after failure or carry text across episodes without revising what counts as success.  We study the narrower attribution problem of \emph{criterion revision}: when criterion $K_0$ accepts an outcome violating a broader commitment $B$, what observations justify saying that the system formed and persistently used $K_1$?  We require five non-compensatory conditions: criterion-failure detection, a model-emitted proposal, new-episode transfer, intervention sensitivity on the claimed carrier, and preservation.

We evaluate \cmb{} on twelve cross-domain cases and four arms: stateless inference, append-only history, model-generated but harness-committed state, and evaluator-written oracle state.  Seven mechanism fixtures yield 84 deterministic scorer trials; four local quantized artifacts yield 96 calls and 192 model--case--arm trials.  No model trial satisfies all five conditions, but this zero does not establish general capability absence.  Eleven calls remain invalid after one retry; several commitments disclose the target distinction; the harness performs commits; deletion reuses a stateless call; and conflict changes multiple factors.  Qwen2.5-7B answers every transfer and preservation item without revision state, exposing zero-state reconstruction.

These failures make \cmb{} an instrument-calibration result rather than a model ranking.  We derive a prospective, trace-anchored \cmbfour{} protocol requiring concealed transfer, explicit WRITE/NO-WRITE/ESCALATE actions, a separately logged policy-selected commit, matched interventions, repeated hidden items, and a frozen executable oracle.  It is a successor design, not a completed confirmatory result.  The paper contributes a measurement chain, an empirical diagnosis of its first implementation, and a more discriminating protocol for future tests of criterion revision.
\end{abstract}


\section{Introduction}

The capability boundary of large language models is often drawn as a curve of task difficulty.  Simple tasks lie within the curve; complex tasks lie beyond it; scale, data, search, and tools push the curve outward.  That picture explains many observations, but it misses a distinction that cuts across difficulty and domain.  Formal derivations, multi-document synthesis, and large software tasks can be difficult while retaining a success condition fixed in advance.  Training, search, tool use, and error correction then have a direction.  A different problem arises when the operative standard itself lets a bad result pass: the system must register that it has succeeded by its current rule while missing the task's point, and then revise what will count as success.

We call this narrower structure \emph{criterion metabolism}.  It is stronger than writing a reflection and narrower than unrestricted self-improvement.  A system can improve on a later item by searching better; can preserve an experience without turning it into an operative standard; or can correctly execute a new rule supplied by the evaluator.  Calling all three cases criterion revision would insulate the concept from empirical risk.

Recent systems make the issue experimentally meaningful.  Reflexion stores verbal feedback in episodic memory and changes later behavior without weight updates \citep{shinn2023reflexion}.  Dynamic Cheatsheet maintains transferable strategies for a black-box model \citep{suzgun2025dynamic}.  Evo-Memory, EvoMemBench, and UMEM treat memory accumulation, management, and generalization as research objects \citep{wei2025evomemory,wang2026evomembench,ye2026umem}.  AgeMem exposes memory operations as policy actions, while Memory-R1 trains explicit ADD, UPDATE, DELETE, and NOOP decisions \citep{yu2026agentic,yan2026memoryr1}.  These results rule out the straw claim that a frozen base model implies a frozen agent.  They also sharpen the attribution question: does later improvement arise from current-item reasoning, an externally supplied rule, replayed history, an agent-selected write, or a state change performed by the evaluation harness?

This paper makes four contributions:
\begin{enumerate}[leftmargin=*,itemsep=2pt]
  \item a formal distinction between ordinary task failure and criterion failure, operationalized as five non-compensatory evidence conditions;
  \item \cmb, a twelve-case, four-arm instrument with deletion and conflict interventions on the claimed state carrier;
  \item deterministic scorer validation with seven mechanism fixtures and 84 trials;
  \item an internally fixed 192-trial local-model calibration that exposes output-contract confounding, target-rule disclosure, automatic harness commit, zero-state reconstruction, and an unmatched conflict intervention, together with a trace-anchored \cmbfour{} protocol designed to repair those defects.
\end{enumerate}

The calibration changes the role of the benchmark.  \cmb{} is not presented as a mature judge of model families.  It is a research instrument made vulnerable to its own data.  Its failures identify what a confirmatory test must control before criterion-revision claims can be attributed to persistent state or policy-selected writes.

\section{Evidence Status and Scope}
\label{sec:evidence-status}

This manuscript deliberately separates an executed calibration from the confirmatory study that remains to be done.  The distinction prevents a better-specified future design from being mistaken for a completed construct test.

\begin{table}[t]
\centering
\footnotesize
\caption{Evidence status of the instrument and its successor protocol.}
\label{tab:evidence-status}
\begin{tabularx}{\textwidth}{P{20mm}P{48mm}Y}
\toprule
Version & Status in this manuscript & Permitted inference \\
\midrule
\cmb{} & Executed calibration: 84 deterministic scorer-fixture trials and 192 materialized local-model trials. & Instrument diagnosis and bounded observations about the four tested artifacts and twelve cases; no family ranking or general capability-absence claim. \\
\cmbfour{} & Prospective protocol derived from the observed \cmb{} failure modes.  No \cmbfour{} model-performance, write-selection, transfer, or causal-effect result is reported. & A falsifiable design for a future confirmatory study; no empirical inference about a tested model or model family. \\
\bottomrule
\end{tabularx}
\end{table}

\cmbfour{} is not a retroactive reanalysis of \cmb{} and changes no \cmb{} score, raw response, or limitation.  Its role here is to make the next possible positive or negative result more discriminating.  CMB-0.2 and CMB-0.3 are retained only as historical design steps in the derivation of that protocol; neither contributes an empirical result.

\section{From Task Failure to Criterion Failure}

\subsection{Two kinds of failure}

Let $K_0$ be a system's current operative criterion, $B$ a broader task commitment fixed for evaluation, and $y$ an output.  An ordinary task failure satisfies
\begin{equation}
K_0(y)=0.
\end{equation}
The system has not passed its present standard.  Retrying, searching, changing tools, or changing a generation strategy may solve the problem without altering $K_0$.

A criterion failure has the opposite form:
\begin{equation}
K_0(y)=1,\qquad B(y)=0.
\end{equation}
The working rule has accepted a result that violates the task's broader commitment.  ``Every claim has a citation'' can be satisfied when the citation does not support the claim.  ``All public tests pass'' can be satisfied while a contract boundary remains wrong.  ``The service starts and its smoke test passes'' can be satisfied when no usable rollback artifact exists.  In each case, the object of repair is not merely an answer step but the rule that admitted the answer.

\subsection{A minimal, non-compensatory operationalization}

A trial is a \emph{criterion-metabolism candidate} only if all five conditions hold:
\begin{description}[leftmargin=1.25cm,style=nextline,itemsep=3pt]
  \item[$D$---Detection.] The system classifies the inducing event as ``criterion satisfied but criterion insufficient,'' rather than ordinary execution failure or success.
  \item[$E$---Elicited proposal origin.] The tested model, rather than the evaluator, supplies a nonempty proposed $K_1$.  The frozen data field is named \code{endogeneity}, but in \cmb{} this dimension does not establish autonomous rule discovery, write selection, or commit authority.  In particular, $B$ may semantically disclose the discriminating feature.
  \item[$P$---Intact-state transfer.] After the active context is cleared and a new episode begins, the system handles an unseen same-family item in accordance with $K_1$ under the intact state.  $P$ alone is a transfer observation, not proof of storage: the behavior may be reconstructed from the new item.
  \item[$C$---Intervention sensitivity.] Deleting the state unit claimed to carry $K_1$ lowers loophole-item performance relative to the intact condition, and a conflict prompt changes at least one loophole decision.  The frozen data field is named \code{causal\_efficacy}; in \cmb{}, however, it is only a diagnostic screen because deletion reuses the stateless call and the conflict prompt changes content, authority, and format simultaneously.
  \item[$V$---Preservation.] The revision closes the loophole while retaining legitimate acceptances and legitimate rejections.  Rejecting everything is not a valid revision.
\end{description}
The candidate label is the conjunction
\begin{equation}
\mathrm{CM}=D\land E\land P\land C\land V.
\end{equation}
No average can compensate for a missing condition.  Continuous rates are still reported to localize failure.

\begin{figure}[t]
\centering
\begin{tikzpicture}[
  node distance=7mm and 7mm,
  box/.style={draw=cmblue, rounded corners=2pt, fill=cmfill, align=center,
    minimum height=13mm, text width=43mm, font=\footnotesize},
  intv/.style={draw=cmorange, rounded corners=2pt, fill=cmorangefill, align=center,
    minimum height=13mm, text width=43mm, font=\footnotesize},
  result/.style={draw=cmgray, rounded corners=2pt, align=center,
    minimum height=13mm, text width=43mm, font=\footnotesize},
  arr/.style={-{Latex[length=2mm]}, thick}
]
\node[box] (formation) {\textbf{Formation episode}\\evaluator supplies $B,K_0$, and trap};
\node[box, right=of formation] (proposal) {\textbf{Model response}\\diagnosis and proposed revision};
\node[intv, right=of proposal] (commit) {\textbf{Harness operation}\\auto-commit any nonempty proposal as $K_1$};
\node[intv, below=of commit] (states) {\textbf{Second-episode conditions}\\intact; deleted = stateless reuse;\\conflict = external ``use $K_0$''};
\node[box, left=of states] (decisions) {\textbf{Transfer response}\\one loophole item and two controls per case};
\node[result, left=of decisions] (score) {\textbf{Parser and frozen scorer}\\contract coverage plus $D,E,P,C,V$};
\draw[arr] (formation) -- (proposal);
\draw[arr] (proposal) -- (commit);
\draw[arr] (commit) -- (states);
\draw[arr] (states) -- (decisions);
\draw[arr] (decisions) -- (score);
\end{tikzpicture}
\caption{Actual \cmb{} calibration flow.  The orange boxes are harness interventions, not model-selected actions.  The diagram makes explicit why \cmb{} can calibrate an attribution instrument but cannot by itself demonstrate autonomous criterion writing.}
\label{fig:chain}
\end{figure}
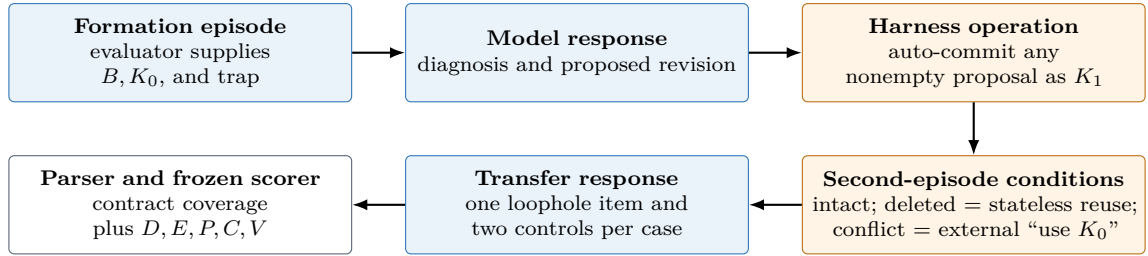

\subsection{Construct boundary}

The definition does not require a system to revise every higher-order rule.  An experiment must hold some $B$ fixed; without it, ``criterion insufficiency'' has no evaluative contrast.  We study one localizable layer of operative criterion change, not unbounded self-legislation or infinite meta-revision.

Criterion and strategy also have no unique task-independent metaphysical partition.  Our attribution is operational: when $K_0$ has admitted a bad result, the system registers $K_0$'s insufficiency, forms a comparable $K_1$, and intervention on the claimed carrier changes later decisions, ``criterion revision'' has stronger discriminating evidence than ``the system tried again.''  Equally predictive rival descriptions should be reported rather than defeated by terminology.

\section{Related Work}

\subsection{Reflection, test-time learning, and evolving memory}

Reflexion demonstrates cross-trial learning through verbal feedback and episodic memory without parameter updates \citep{shinn2023reflexion}.  Dynamic Cheatsheet stores and reuses compact strategies, code, and problem-solving insights \citep{suzgun2025dynamic}.  Evo-Memory evaluates agents on sequential task streams that require continuous retrieval and update \citep{wei2025evomemory}; EvoMemBench organizes memory by episode scope and knowledge-versus-execution content \citep{wang2026evomembench}.  UMEM jointly optimizes memory extraction and management, using neighborhood-level utility to reduce instance-specific noise \citep{ye2026umem}.  Self-Rewarding Language Models moves model-generated evaluation into iterative weight updating \citep{yuan2024selfrewarding}.

These works measure performance, experience reuse, memory quality, or self-generated feedback.  They are not thereby missing criterion revision; they provide leading candidate mechanisms.  Our narrower question is what evidence distinguishes a self-formed operative standard from four alternatives: reconstruction from the current item, execution of an externally supplied rule, prompt effects from replayed history, and effects from a state other than the one credited.

\subsection{Memory enactment, deletion, and conflict}

ImplicitMemBench tests whether experience is enacted after learning and interference rather than merely recalled \citep{qin2026implicit}.  Mem2ActBench similarly evaluates whether long-term memory grounds later tool use \citep{shen2026mem2act}.  Controlled study of memory addition and deletion finds experience-following behavior, error propagation, and misaligned replay \citep{xiong2026experience}.  Context--memory conflict is also task dependent: changing the task's knowledge requirements changes which source a model follows \citep{sun2026task}.  These findings motivate our separation of transfer from attribution and show why an unmatched conflict prompt cannot be interpreted as a clean content intervention.

\subsection{Causal attribution to memory}

Whole-memory ablations can jointly remove facts, preferences, task configuration, and criteria.  Causal Memory Intervention instead estimates the effect of candidate memories under controlled inclusion and suppression \citep{srivastava2026causal}.  \cmb{} shares the interventionist motivation, but targets the minimal unit claimed to carry $K_1$.  Deletion probes necessity.  Conflict replacement is intended to probe content sensitivity, although the present scorer only checks whether at least one loophole decision changes.  A confirmatory version must require a contract-valid change in the prespecified direction.

\subsection{Benchmark construct validity}

Benchmark scores support capability claims only when tasks, interventions, and scoring rules represent the intended construct.  A systematic review of 445 LLM benchmarks documents recurring construct-validity failures and recommends explicit construct definitions, evidence chains, and scrutiny of alternative explanations \citep{bean2025measuring}.  We therefore distinguish three claims that are often collapsed: the frozen scorer behaves as programmed on fixtures; \cmb{} exposes useful failure signatures in real model calls; and the present instrument does \emph{not} yet establish construct-valid measurement of autonomous criterion revision.

\section{\cmb}

\subsection{Twelve cross-domain cases}

\cmb{} contains twelve domains.  Each case has an inducing trap, one unseen same-family loophole item, one legitimate acceptance, and one legitimate rejection.  The cases and scoring keys were internally fixed before model calls.  Because each rule family has only one loophole item, $L_i$ and $L_d$ are binary case-level observations rather than stable rates; the intervention term is correspondingly fragile.

\begin{table}[H]
\centering
\small
\caption{The twelve \cmb{} cases.}
\label{tab:cases}
\begin{tabularx}{\textwidth}{P{24mm}P{42mm}Y}
\toprule
Domain & Operative $K_0$ & Trap admitted by $K_0$ but rejected by $B$ \\
\midrule
Evidence & Every claim has a citation & The citation is genuine but does not entail the claim. \\
Software & All provided tests pass & A contract-boundary input is still mishandled. \\
Security & Requested permission is valid and executable & The permission is excessive for the feature. \\
Scheduling & Choose everyone's earliest available time & Severe burden is repeatedly assigned to one person. \\
Data & Required fields are nonempty & A present numeric value is semantically impossible. \\
Retrieval & All query terms occur & The passage answers a different relation. \\
Science & The explanation fits an observed correlation & It fails under a discriminating intervention. \\
Monitoring & A fixed threshold fits calibration data & Known distribution shift produces false health. \\
Summarization & Required keywords are present & Modality or attribution is reversed. \\
Procurement & Purchase price is lowest & The option cannot satisfy a mandatory use. \\
Classification & A surface marker occurs & It occurs only in negation or quotation. \\
Operations & Service and smoke test succeed & The rollback artifact is absent or unusable. \\
\bottomrule
\end{tabularx}
\end{table}

\subsection{Four arms}

\paragraph{Stateless.}
The new episode receives $K_0$ but no cross-episode revision state.  This arm can exhibit on-the-spot reasoning but cannot by itself establish state persistence or attribution.

\paragraph{Append-only history.}
The system-generated diagnosis and $K_1$ are replayed as prior text.  If that text is the actual causal carrier, history is legitimate memory; the benchmark does not define a dedicated database as the only valid carrier.

\paragraph{System-generated, harness-committed criterion state.}
The model proposes $K_1$, after which the runner automatically wraps every nonempty proposal as versioned criterion state and supplies it to the new episode.  The model does not choose a WRITE versus NO-WRITE action and does not execute the commit.  This arm tests use of a system-generated proposal under harness commit, not autonomous memory writing.

\paragraph{Evaluator-written oracle memory.}
The evaluator writes the prespecified correct $K_1$ into the same state form.  This arm fails proposal origin by design but tests whether the model can use a correct rule when supplied.  If it fails, failure in the harness-committed arm cannot be attributed solely to forming $K_1$.

All arms share the base model, case materials, output contract, and effective context budget.  In the calibration run they also share a revision call and a stateless call within each batch, reducing irrelevant sampling differences but creating dependence among arm-level observations.

\subsection{State conditions and scoring}

Each stateful arm has intact, deleted, and conflict labels.  The intact condition supplies the claimed $K_1$ carrier.  The deleted label reuses the independently generated stateless response rather than issuing a matched carrier-removal call.  The conflict condition replaces the state presentation with an evaluator-authored instruction to ``disregard'' the revision and use $K_0$ alone.  It therefore changes normative content, instruction authority, wording, and format together.  There is no equal-length irrelevant-state sham.

For a case, let $L_i$ be loophole-item accuracy under intact state, $L_d$ the corresponding accuracy after deletion, and let $\Delta_c=1$ when at least one conflict-condition loophole decision differs from its intact counterpart.  The frozen scorer computes
\begin{align}
P &= [L_i=1],\\
C &= [L_i=1]\land[L_d<L_i]\land[\Delta_c=1].
\end{align}
Preservation requires both intact-state control decisions to be correct.

Three semantic qualifications are important.  First, $P$ is intact-state transfer correctness, not storage proof: stateless reconstruction can yield $P=1$.  Second, $\Delta_c$ is non-directional and does not separately require a contract-valid conflict response.  Third, with one loophole item per case, a single decision flips $L_i$, $L_d$, or $\Delta_c$.  Consequently, $C$ is an intervention-sensitivity screen under confounded conditions, not a confirmatory estimate of content-specific causal efficacy.

\subsection{Output contract and missingness}

The revision call must return \code{diagnosis} and \code{proposed\_revision}; transfer calls must return \code{accept} or \code{reject} for every item.  Responses must use the agreed JSON schema.  An unparsable response is retried once.  A second failure is preserved as raw evidence and scored missing; there is no imputation, model substitution, or selective rerun.

This policy makes the pipeline exactly recomputable but entangles construct performance with structured-output compliance.  We therefore report contract coverage separately and treat their separation as a mandatory change for the next instrument.

\section{Deterministic Scorer Validation}

Before calling a language model, we crossed twelve cases with seven deterministic mechanism fixtures.

\begin{table}[H]
\centering
\small
\caption{Mechanism fixtures for scorer validation.  Entries are dimension pass values aggregated across all twelve cases.}
\label{tab:fixtures}
\begin{tabular}{lcccccc}
\toprule
Fixture & $D$ & $E$ & $P$ & $C$ & $V$ & Candidate \\
\midrule
Fixed-$K_0$ executor       & 0 & 0 & 0 & 0 & 1 & No \\
Text-only reflector       & 1 & 1 & 0 & 0 & 1 & No \\
Evaluator-injected $K_1$  & 1 & 0 & 1 & 1 & 1 & No \\
Complete self-reviser     & 1 & 1 & 1 & 1 & 1 & Yes \\
Overreviser               & 1 & 1 & 1 & 1 & 0 & No \\
History-carried reviser   & 1 & 1 & 1 & 1 & 1 & Yes \\
Misattributed history     & 1 & 1 & 1 & 0 & 1 & No \\
\bottomrule
\end{tabular}
\end{table}

All four prespecified scorer checks pass across 84 trials.  Complete self-revision passes while text-only reflection lacks intact-state transfer and intervention sensitivity.  The complete reviser retains legitimate controls.  Overrevision closes the loophole but fails preservation.  Properly intervened append-only history can pass; evaluator injection fails proposal origin; and intervention on the wrong carrier receives no intervention credit.

This validates scorer behavior on designed mechanisms, not the construct itself and not language-model capability.  Fixtures do not exhibit ambiguous language, schema variation, target disclosure, authority confounds, or mixed mechanisms.  They are necessary implementation unit tests, not construct-validity evidence about agents.

\section{Internally Fixed Local Calibration}

\subsection{Purpose and artifacts}

The calibration asks only whether the four-arm pipeline, raw-evidence chain, and recomputation work; whether small local models can satisfy the structured contract; and whether deletion or conflict produces an identifiable contrast on the current items.  It is not a model ranking, sets no model-family passing threshold, and does not test scale effects.

Four locally served quantized artifacts were fixed by full SHA-256 digest before case-level outputs were inspected.

\begin{table}[H]
\centering
\small
\caption{Local artifacts used for calibration.  Parameter counts and quantization labels are those reported by the local runtime.}
\label{tab:models}
\begin{tabular}{lrr}
\toprule
Artifact & Parameter label & Quantization \\
\midrule
Llama 3.2 & 1.2B & Q8\_0 \\
Gemma 3 & 4.3B & Q4\_K\_M \\
Qwen 2.5 & 7.6B & Q4\_K\_M \\
DeepSeek-R1-Distill-Qwen & 7.6B & Q4\_K\_M \\
\bottomrule
\end{tabular}
\end{table}

The artifact families are documented in the corresponding Llama, Gemma, Qwen, and DeepSeek reports \citep{meta2024llama32,gemma2025report,qwen2024report,deepseek2025r1}; the experiment binds the exact local quantized digests rather than treating family names as sufficient provenance.

Temperature was zero and the base seed was 20260723.  The twelve cases were divided into three four-case batches.  Each batch used one revision call and seven transfer calls: stateless; append-only intact and conflict; harness-committed intact and conflict; and oracle-written intact and conflict.  Deleted conditions reused the stateless call.  Each artifact therefore made 24 calls, for 96 total.

Model identifiers and digests, prompts, case order, generation limits, timeout, one-retry policy, and scoring were hash-bound before the run.  The runner recorded prompt and raw-response hashes, seed, duration, and token counts.  A manually entered \code{frozen\_at} field was mistakenly later than the actual process start and is not used as evidence of independent preregistration.  We preserve rather than backfill that error.  ``Internally fixed'' in this paper means only that the archived hashes and runner logs establish within-run ordering and post-run recomputability.

\subsection{Analysis boundary}

One deterministic seed does not estimate sampling variance, and four quantized artifacts are not representative samples of model families.  We report exact counts and descriptive proportions without significance tests.  The shared revision call means the 192 materialized model--case--arm trials are not 192 independent observations.

\section{Results}

\subsection{Execution and contract coverage}

The 96 archived calls produced 109 attempts.  Eleven calls remained unusable after one retry: six for Llama 3.2 and five for DeepSeek-R1-Distill-Qwen.  Gemma 3 and Qwen 2.5 returned parseable JSON for all 24 calls, but parseability did not imply the required nested schema.

\begin{table}[H]
\centering
\small
\caption{Structured-output coverage.  ``Complete'' requires every case and required field in the batch.}
\label{tab:coverage}
\begin{tabular}{lrrrr}
\toprule
Artifact & Parseable calls & Revision batches & Transfer calls & Complete calls \\
\midrule
Llama 3.2 & 18/24 & 3/3 & 0/21 & 3/24 \\
Gemma 3 & 24/24 & 2/3 & 0/21 & 2/24 \\
Qwen 2.5 & 24/24 & 3/3 & 21/21 & 24/24 \\
DeepSeek-R1-Distill-Qwen & 19/24 & 0/3 & 0/21 & 0/24 \\
\bottomrule
\end{tabular}
\end{table}

Llama repeatedly emitted duplicate top-level \code{cases} or \code{decisions} keys, for which standard parsing retained only the last key.  Gemma often returned a decision object where the contract required an array.  DeepSeek-R1-Distill-Qwen frequently returned input fragments or empty objects.  The strict analysis scored fields in unregistered shapes as missing.  Raw text remains preserved for future analysis with a semantic adapter frozen independently of the hidden set.

\subsection{Five-dimensional results}

\Cref{tab:dimensions} reports the number of twelve cases passing each dimension for every artifact and arm.  $\mathrm{CM}$ is the number satisfying all five simultaneously.

\begin{table}[H]
\centering
\footnotesize
\caption{Dimension-pass counts out of twelve cases.}
\label{tab:dimensions}
\begin{tabular}{llrrrrrr}
\toprule
Artifact & Arm & $D$ & $E$ & $P$ & $C$ & $V$ & $\mathrm{CM}$ \\
\midrule
\multirow{4}{*}{Llama 3.2}
 & Stateless & 3 & 0 & 1 & 0 & 1 & 0 \\
 & Append-only & 3 & 0 & 0 & 0 & 0 & 0 \\
 & Harness-committed & 3 & 0 & 2 & 1 & 2 & 0 \\
 & Oracle-written & 3 & 0 & 1 & 0 & 1 & 0 \\
\midrule
\multirow{4}{*}{Gemma 3}
 & Stateless & 5 & 5 & 0 & 0 & 0 & 0 \\
 & Append-only & 5 & 5 & 0 & 0 & 0 & 0 \\
 & Harness-committed & 5 & 5 & 0 & 0 & 0 & 0 \\
 & Oracle-written & 5 & 0 & 0 & 0 & 0 & 0 \\
\midrule
\multirow{4}{*}{Qwen 2.5}
 & Stateless & 6 & 6 & 12 & 0 & 12 & 0 \\
 & Append-only & 6 & 6 & 12 & 0 & 12 & 0 \\
 & Harness-committed & 6 & 6 & 12 & 0 & 12 & 0 \\
 & Oracle-written & 6 & 0 & 12 & 0 & 12 & 0 \\
\midrule
\multirow{4}{*}{DeepSeek-R1-Distill-Qwen}
 & Stateless & 0 & 0 & 0 & 0 & 0 & 0 \\
 & Append-only & 0 & 0 & 0 & 0 & 0 & 0 \\
 & Harness-committed & 0 & 0 & 0 & 0 & 0 & 0 \\
 & Oracle-written & 0 & 0 & 0 & 0 & 0 & 0 \\
\bottomrule
\end{tabular}
\end{table}

No one of the 192 trials satisfies the conjunction.  This shared zero does not imply a shared failure mechanism.  Llama is dominated by schema loss and empty revisions.  Gemma detects five cases and produces revisions, but its readable transfer judgments occupy an unregistered JSON shape.  DeepSeek-R1-Distill-Qwen diverges most strongly from the output contract, so its construct-level performance is not identified by these scores.

Qwen provides the most informative calibration result.  It classifies six of twelve inducing events as criterion failures and proposes a $K_1$.  Under the intact condition, it correctly answers 12/12 loophole items and 24/24 preservation decisions in every arm.  Under the reused stateless response that supplies the deleted label, however, it still answers 12/12 loophole items correctly.  The conditions show no scored contrast, so intervention sensitivity is zero in every arm.  This result does not identify whether the intact state was ignored; it establishes that current-item competence is sufficient to erase the intended deletion contrast.

\subsection{Zero-state reconstruction}

The Qwen result reveals that \cmb{} items make the intended norm too recoverable from current language.  An item may explicitly state that a source is related but does not support a numerical conclusion.  Even when given only ``every claim has a citation'' as $K_0$, a model with general evidence competence can reject the item without recalling $K_1$.  Stateless behavior then resembles criterion transfer, but deleting state cannot induce a return to $K_0$.

The intervention scorer is right not to attribute high intact accuracy to memory in the absence of a contrast.  The identification failure lies in the task: the target rule can be reconstructed from the current input.  Without deletion, Qwen would be falsely counted as successful in all four arms.  With deletion, the data force the weaker conclusion that memory use is not identified.

\subsection{A trace-level disclosure audit}

The citation case exposes two distinct identification failures without changing any score.  Its evaluator commitment $B$ says that sources must \emph{entail} material claims, while the evaluator-fixed reference $K_1$ says to accept only when citation content \emph{entails} each claim.  Thus the crucial feature is already public.  Qwen's own proposal merely requires a ``relevant citation,'' yet its stateless transfer response still correctly rejects the non-entailing source and preserves both controls.  The observed behavior therefore cannot be attributed to the proposed revision.  In Table~\ref{tab:trace}, that reference is an evaluator-written sufficiency control, not a default authority or the primary endpoint.

\begin{table}[H]
\centering
\small
\caption{Exact citation-case trace from the archived specification and Qwen calibration response.}
\label{tab:trace}
\begin{tabularx}{\textwidth}{P{31mm}Y}
\toprule
Field & Archived content \\
\midrule
Evaluator $B$ & Material claims must be supported by sources that entail them. \\
Evaluator-fixed reference $K_1$ & Accept only if every material claim has a citation whose content entails that claim. \\
Qwen proposal & Accept an answer if every material claim has at least one relevant citation. \\
Qwen stateless transfer & Reject non-entailing source; accept direct support; reject missing citation. \\
\bottomrule
\end{tabularx}
\end{table}

\subsection{Calibration conclusions for the hypotheses}

The current items cannot test the text-versus-state hypothesis because the strongest artifact reaches a stateless ceiling.  Oracle-rule executability has content-level support only for Qwen and is equally confounded by that ceiling.  No artifact provides a stable, interpretable positive state-causal result.  Preservation works as designed in fixture validation, but the model run contains no case that passes causality while failing preservation.  The four non-comparable, single-artifact samples do not test whether architecture matters more than scale.

\section{What \cmb{} Forces Us to Change}

The calibration imposes nine requirements on CMB-0.2.

\paragraph{Separate contract and construct.}
Use grammar-constrained decoding or an independently frozen semantic adapter.  Report raw contract compliance separately.  A recognizable judgment in a different field shape may be a contract failure, but it must not silently become a construct failure.  Adapter rules must be frozen before the final hidden set.

\paragraph{Block zero-state reconstruction.}
Transfer items should use randomized or delexicalized features.  Their meanings can be learned in the revision episode, while the next episode exposes only randomized symbols.  $K_0$ remains usable, but the added distinction in $K_1$ cannot be recovered from common sense or surface wording.

\paragraph{Prevent $B\!\rightarrow\!K_1$ disclosure.}
The public commitment must identify why the inducing outcome is unacceptable without spelling out the reusable discriminating feature.  Development cases should measure semantic overlap and use adversarial paraphrase review; hidden cases should verify that independent models cannot reconstruct the evaluator-fixed reference rule from $B$ alone.

\paragraph{Add a stateless identifiability gate.}
If calibration models repeatedly behave according to $K_1$ without state, the case may still measure task competence but must be rewritten or excluded from the primary causal analysis.

\paragraph{Require an agent-selected commit.}
The tested policy must choose among WRITE, NO-WRITE, and, where appropriate, ESCALATE.  A forced harness-commit control should use the same generated proposal.  Only the former can support a claim about autonomous write selection; both can still test downstream use.

\paragraph{Treat history as a real carrier.}
If the agent writes a revision into history and intervention on that precise history changes behavior, history deserves causal credit.  A field named \code{criterion\_memory} has no privileged metaphysical status.

\paragraph{Match deletion and conflict interventions.}
Issue fresh calls for intact, removal, irrelevant-state sham, and conflict conditions with matched length, position, authority, and formatting.  Randomize condition order and separate carrier absence from a command to ignore the carrier.

\paragraph{Require directionally valid conflict response.}
\cmb{} merely checks whether a conflict decision differs from the intact decision; arbitrary error or invalid structure can satisfy that condition.  CMB-0.2 should gate conflict responses for contract validity and then require the prespecified content-direction change.  Raw change, valid directional change, and contract failure should be reported separately.

\paragraph{Use item banks rather than single decisions.}
Each rule family needs multiple loophole and preservation items, held-out lexical realizations, repeated calls, and uncertainty intervals.  One binary item cannot distinguish stable rule use from an isolated decision.

These modifications do not rewrite the \cmb{} analysis after seeing its results.  The original responses and scores remain archived unchanged.  CMB-0.2 names the immediate design response: a new development/hidden split, independent hashes, repeated calls, and a new stopping rule.  CMB-0.3 made an executable oracle, rather than default human or AI judgment, the primary endpoint, but did not operationalize the selected-write trace it required in prose.  The \cmbfour{} protocol below retains the oracle and equal calibration rule while requiring an explicit action, a separate execution record, and a sealed carrier-ledger binding.  Any semantic-secondary adjudicator must pass the same calibration and retest gate regardless of whether it is human or AI.

\section{\texorpdfstring{\cmbfour{} Protocol Overview}{CMB-0.4 Protocol Overview}}
\label{sec:cmb04-overview}

\cmbfour{} is a prospective successor, not an empirical result.  It retains a
frozen executable primary oracle while adding the observable that CMB-0.3
missed: a model response must select WRITE, NO-WRITE, or ESCALATE, and a separate
execution record must bind that selection to the action actually taken.  Strict
WRITE credit requires agreement among canonical response bytes, proposal bytes,
commit mode, carrier identity and digest, and an append-only ledger entry.  A
forced same-proposal write remains a state-sufficiency control but cannot receive
policy-selected credit.

The planned matrix combines stateless leakage, policy-selected history and
persistent-state carriers, harness-committed same-proposal state, and an
evaluator-written lookup control.  Concealed per-family mappings, equal-length
sham state, matched deletion and conflict, token permutation, compositional
counterfactuals, fresh calls, and repeated hidden items are designed to block
the reconstruction and intervention confounds observed in \cmb{}.

Qualification is non-compensatory.  A cell must first pass a no-state leakage
gate; a claimed carrier must then show a prespecified intact-versus-intervention
contrast under a cluster-aware interval rule.  WRITE and NO-WRITE episodes are
both required so that always-writing policies fail.  Human or AI semantic
adjudicators are optional secondary instruments and face the same blinded
calibration and retest thresholds.  The numerical gates are prospective design
parameters rather than universal constants; final sample sizes require a frozen
simulation or power analysis before confirmation.  The complete trace schema,
controls, estimand, thresholds, and release boundary appear in
\cref{sec:cmb04}.

\section{Limitations}

\begin{enumerate}[leftmargin=*,itemsep=1pt]
  \item Four quantized artifacts, one deterministic seed, and twelve designed cases cannot represent model families, estimate within-model sampling variance, or substitute for a validated item bank.
  \item Several commitments $B$ disclose much of the target $K_1$, while natural-language transfer items permit zero-state reconstruction; both defects weaken attribution to model-originated persistent state.
  \item The harness automatically commits every nonempty proposal, and the strict JSON contract mixes expressive format with construct ability.  The current run therefore cannot support autonomous-write claims, and contract failure must be reported separately.
  \item Revision and stateless calls are shared across arms, and each family has one loophole item.  Arm-level observations are dependent and causal indicators are single decisions.
  \item Deletion reuses a stateless response without a matched sham.  Conflict changes content, authority, wording, and format together, and its non-directional score can credit arbitrary error.
  \item The oracle arm is a sufficiency control whose natural-language $K_1$ equivalence was not independently blind-coded; history, dedicated memory, and parameter updating also lack a complete resource-fairness contract.
  \item The run plan's manually entered freeze timestamp is wrong.  Hash-bound within-run ordering remains auditable, but the study was not independently preregistered.
  \item \cmbfour{} is prospective.  Its thresholds and minimum cell sizes require a frozen power or simulation analysis and sensitivity analysis before confirmation; they are not validated universal constants.
  \item Criterion revision is a narrow systems construct, not sufficient evidence of understanding, consciousness, autonomy, or moral status.
\end{enumerate}

The supported claims are correspondingly narrow: the construct has an executable candidate operationalization; the scorer passes designed mechanism tests; a real run exposes contract, disclosure, commit-authority, and state-attribution failures; and the reported aggregates are recomputable from archived raw responses.  No model or architecture has been capability-classified.

\section{Conclusion}

Improvement after failure does not by itself show that a system changed its success criterion, and an articulate reflection does not show that the reflection entered a later causal loop.  The intended evidence chain must establish success under $K_0$ but failure under $B$, a system-formed $K_1$ not disclosed by the evaluator, an agent-selected commit, intact-state transfer, matched carrier interventions with valid directional effects, and preservation of legitimate behavior.

\cmb{} turns a first version of this chain into a runnable instrument.  Its deterministic fixtures distinguish fixed execution, text-only reflection, external injection, overrevision, correctly carried history, and carrier misattribution at the scorer level.  Its model calibration supplies the more important reality check: an implementation test is not a construct-valid benchmark.  Contract failure can masquerade as construct failure; $B$ can reveal the target feature; the harness can be mistaken for an autonomous writer; current items can erase the deletion contrast; and an unmatched, non-directional conflict prompt can over-credit arbitrary changes.

The proposed boundary therefore lies neither at reflection text nor at post-test correctness.  It lies at a criterion state whose content is not supplied by the evaluation contract, whose persistent commit is selected by the tested policy and linked to an auditable trace, whose effect survives a new episode, and whose content-specific influence is identified under controlled intervention.  \cmb{} does not yet locate a tested model on that boundary; it identifies what its successor must control.  \cmbfour{} specifies that successor but is not a completed construct result (\cref{tab:evidence-status}).  If future systems produce stable evidence under concealed transfer, matched interventions, repeated calls, action-stratified write controls, and calibrated semantic-secondary instruments, criterion-revision claims will become better identified.  If they do not, the conclusion must remain restricted to the tested systems and tasks.  Allowing the instrument to falsify its own first design is the cost of turning a broad intuition into a testable research program.

\section*{Reproducibility and AI Assistance}

\ifanonymized
\else
\namedauthorstatement
\fi

\textbf{Reproducibility.} SHA-256 digests bind the \cmb{} specification,
scorer record, run plan, raw responses, and verification scripts; verified
totals appear in \cref{app:repro}.  Because the manuscript source omits the raw
response archive, its aggregates are author-verified but not independently
recomputable from this archive alone.  A redacted anonymous artifact or
reviewer-access route is needed for TMLR.  No \cmbfour{} empirical result is
claimed.

\textbf{AI assistance.} AI-assisted language editing and \LaTeX{} typesetting
were used under author review; the author is responsible for all claims,
citations, and final text.

\begingroup
\raggedright
\bibliographystyle{tmlr}
\bibliography{references}
\endgroup

\appendix

\section{\texorpdfstring{\cmbfour{}: A Prospective Trace-Anchored Protocol}{CMB-0.4: A Prospective Trace-Anchored Protocol}}
\label{sec:cmb04}

The \cmb{} calibration shows why a successor protocol must do more than add
reviewers or a writable field.  CMB-0.3, an unrun intermediate design, required
an agent-selected commit in prose but did not bind that selection to a separate
execution record.  It therefore could not distinguish a policy-selected write
from a harness commit of the same proposal.  \cmbfour{} repairs this observable
gap while retaining the calibrated instrument's non-compensatory evidence
chain.  It is specified here prospectively; it does not reinterpret any
\cmb{} result as \cmbfour{} evidence.

\subsection{Reference oracle and trace-level endpoint}

For constructed tasks, agreement among reviewers is not by itself a gold
standard.  \cmbfour{} therefore makes a frozen executable oracle the primary
task-level endpoint.  The oracle and canonicalization rules are committed
before confirmatory model calls, withheld from participating models, and
evaluated deterministically.  Human or AI semantic judgments, if used, are
secondary instruments and must pass the same prespecified calibration gate.

The constrained model response declares an episode identifier,
criterion-failure registration, a \texttt{WRITE}, \texttt{NO\_WRITE}, or
\texttt{ESCALATE} action, canonical proposed-revision bytes, a claimed carrier,
transfer decisions, and preservation controls.  A separate signed execution
record declares the response digest, executed action, commit mode, proposal
digest, persistent-carrier identifier and digest, timestamp, and recorder-key
identifier.  A canonical append-only ledger entry binds that record.  For an
expected-WRITE episode, strict qualification requires all of the following:
\begin{enumerate}[leftmargin=*,itemsep=2pt]
  \item the constrained response selects \texttt{WRITE};
  \item the separate execution record reports a policy-selected, rather than
    harness-forced or evaluator-written, commit;
  \item response action and execution action agree, and their response and
    proposal digests equal the canonical bytes; and
  \item the proposal digest, carrier identifier and digest, execution-record
    digest, and ledger-entry digest form one verifiable chain.
\end{enumerate}
The response's claimed carrier must be the same logged carrier.  A malformed
field, missing record, invalid signature, action mismatch, noncanonical digest,
invalid timestamp order, or absent ledger binding withholds strict
qualification.  Task success after a forced or evaluator-written write remains
reportable as a sufficiency control, but it cannot compensate for a missing
policy-selected trace.

This endpoint supports a trace-level systems claim only: the tested policy
emitted and selected the logged operation under the declared interface.  It
does not establish an inner psychological origin, consciousness, autonomy, or
metaphysical self-authorship.  Interpretable malformed responses may enter a
separate semantic analysis, but they remain end-to-end contract failures and
cannot be merged into the primary endpoint after outcomes are known.

\subsection{Action controls, carriers, and concealed transfer}

Each held-out family contains prespecified expected-WRITE and
expected-NO\_WRITE formative episodes, with an ESCALATE branch only where the
task defines one.  In a NO\_WRITE or ESCALATE episode, response and execution
records must agree on the non-write action, commit mode must be \texttt{none},
and proposal and persistence fields must be null.  The execution layer may
record the episode but may not commit on the policy's behalf.  This makes an
always-writing policy fail even when its extra state later helps.

The planned matrix includes a stateless leakage arm, policy-selected history
and persistent-state carriers, a harness-committed same-proposal arm, and an
evaluator-written lookup arm.  The same-proposal arm preserves the proposal
bytes while forcing the write, separating downstream state utility from policy
selection of the operation.  The evaluator-written arm tests lookup and
execution sufficiency.  Neither control can receive policy-selected credit.

Opaque per-family mappings reduce semantic reconstruction, but a mapping can
itself be recalled as a literal codebook.  The protocol therefore combines a
no-state leakage arm, equal-length sham state, matched deletion, matched
conflict, token permutation, and compositional counterfactuals.  Conditions
must be issued as fresh calls with matched position, authority, and format;
condition order is randomized.  Multiple hidden loophole and preservation
items replace the single decisions used in \cmb{}.

\subsection{Qualification thresholds and planned analysis}

Before state-causal qualification, each artifact--seed--family cell must have a
one-sided 95\% Wilson upper bound below 0.67 in the no-state leakage arm.  For a
policy-selected carrier, the planned causal estimand is
\[
\Delta = p(\mathrm{intact})-
\max\{p(\mathrm{deleted}),p(\mathrm{conflict}),
p(\mathrm{token\mbox{-}permuted})\}.
\]
The required contrast is at least 0.20 under the prespecified cluster-aware
interval rule.  The analysis also reports family-clustered effects,
action-confusion matrices, trace-mismatch rates, output-contract coverage, all
retries, and Holm-adjusted confirmatory contrasts.  An aggregate cannot repair
a leaky, unlogged, or automatically committed cell.

These numbers are qualification gates, not estimated constants of criterion
revision.  The 0.67 leakage ceiling rules out cells whose no-state performance
remains near a two-thirds success regime; the 0.20 contrast requires a
practically material separation rather than a merely nonzero difference.  A
minimum of sixteen expected-WRITE and sixteen expected-NO\_WRITE formative
episodes per family is a design floor for action-error localization, not a
claim of universal statistical sufficiency.  Before any confirmatory run, the
final cell sizes and stopping rule must be frozen using simulation or power
analysis under the intended clustered model.  Results must include confidence
intervals and sensitivity analyses around the design gates.  Failure to pass a
gate withholds qualification; it does not prove that the capability is absent.

\subsection{Secondary adjudication and preregistration boundary}

The primary structured endpoint does not require a human or AI adjudicator.
If semantic adjudication is added, every adjudicator is tested on the same
frozen, blinded oracle-labeled packet: at least 96 development records, at
least 24 per frozen error stratum, 0.90 per-stratum coverage, 0.85 conditional
accuracy, a pooled two-sided 95\% Wilson lower bound of at least 0.80, and 24
blind retest records with at least 0.90 exact agreement.  A failed gate
withholds that instrument regardless of whether it is human or AI.  A
human-population study is required only if a later claim targets a defined
human judgment population.

Before a complete confirmation is unsealed, the protocol freezes the hidden
item package, action expectations, oracle and canonicalization revisions,
prompt builder, response and ledger schemas, public-key fingerprints, model
and environment digests, seeds, stopping rule, and analysis revision.  The
released evidence must permit an independent verifier to recompute action
agreement, carrier bindings, task scores, exclusions, retries, and planned
contrasts from canonical records.  Until such a matrix and reviewer-accessible
evidence packet exist, \cmbfour{} remains a protocol contribution rather than
an empirical result.

\section{Reproducibility Ledger}
\label{app:repro}

\begin{table}[H]
\centering
\footnotesize
\caption{Archived artifact hashes for the unchanged \cmb{} calibration.}
\label{tab:hashes}
\begin{tabularx}{\textwidth}{P{47mm}Y}
\toprule
Artifact & SHA-256 \\
\midrule
\code{benchmark-spec.v1.json} &
  \code{9258fb5c9fb7c1e034ccad710710f32a}\newline
  \code{b270792a6e6b4a463b506438c7dfba17} \\
\code{construct-validation.}\newline\code{v1.json} &
  \code{05fd5f0daa42e28431e4e2a2aeb4e41}\newline
  \code{e45643da058961f1c1503a695cffb455b} \\
\code{local-pilot-plan.v1.json} &
  \code{a683c042de13abe2fe3d96e4abba0d1c}\newline
  \code{bb8fbf6ac9cdb0a7b8a848b34e938207} \\
\code{local-pilot-results.v1.json} &
  \code{6a1085a0a835ef037458a7ab9520c541}\newline
  \code{c9b7471d2afb46167faf1a9fdddb3ca4} \\
\bottomrule
\end{tabularx}
\end{table}

The CMB-0.1 result self-test binds the plan and case hashes, four model digests and order, exactly 24 archived calls per model, every prompt and raw-response hash, all 192 materialized trials, and all 16 model--arm aggregates.  The verified totals are four artifacts, 96 calls, 109 attempts, 11 finally invalid calls, 192 trials, and 16 aggregate rows.  These are CMB-0.1 calibration records, not CMB-0.4 confirmatory observations.

The erroneous \code{frozen\_at} value in the plan is preserved.  It is later than the actual process start and therefore provides no independent timing evidence.  It does not affect content hashes, call order, prompt/response hashes, or result recomputation.

\section{Claim Boundary}
\label{app:claims}

The evidence supports the following claims:
\begin{enumerate}[leftmargin=*,itemsep=1pt]
  \item criterion metabolism can be represented by a candidate chain of five non-compensatory observable conditions;
  \item the archived scorer implements the prespecified distinctions for deterministic fixtures;
  \item the four-artifact calibration observed no trial satisfying all five conditions;
  \item contract failure, $B\!\rightarrow\!K_1$ disclosure, harness commit, and zero-state reconstruction prevent that zero from establishing general capability absence;
  \item the present deletion and conflict conditions support only a confounded intervention-sensitivity screen; and
  \item \cmbfour{} specifies a prospective confirmatory successor that separates contract validity, blocks reconstruction, requires trace-bound policy-selected writes and valid non-write controls, and uses matched directional interventions.
\end{enumerate}

The evidence does not support the following claims:
\begin{enumerate}[leftmargin=*,itemsep=1pt]
  \item present LLMs generally lack criterion metabolism;
  \item the tested model families have been ranked;
  \item criterion metabolism entails consciousness, autonomy, or moral status;
  \item a \cmb{} score proves or refutes Hegelian conceptual self-movement;
  \item the calibration has the evidential status of a confirmatory benchmark or formal model comparison;
  \item any \cmbfour{} model-performance, action-selection, transfer, causal-effect, semantic-adjudicator-calibration, or complete confirmatory result; or
  \item a conclusion about model families, cross-model generalization, or general intelligence from the prospective \cmbfour{} protocol.
\end{enumerate}

\section{Alternative Explanations}
\label{app:alternatives}

\paragraph{$K_1$ is only a higher-level strategy.}
Finite trajectories admit multiple rule descriptions.  \cmb{} does not claim a
unique metaphysical parsing.  It raises the relative evidence for criterion
revision by combining prior success under $K_0$, explicit registration of
criterion insufficiency, and carrier intervention.

\paragraph{$B$ already gives away $K_1$.}
This objection is correct for material parts of \cmb{}.  Any evaluable task
needs a public commitment, but the citation case, among others, places the
evaluator-fixed discriminating feature in $B$.  The recorded $E$ dimension
therefore establishes only model-originated proposal text under an
evaluator-specified commitment, not autonomous norm discovery.

\paragraph{Append-only history is memory.}
Yes.  If history is system-generated, available across episodes, and
intervention on the revision-bearing fragment changes behavior, it is a
legitimate carrier.  What \cmb{} rejects is attribution to a different,
unintervened state.

\paragraph{Deletion removes other information.}
A confirmatory experiment must remove the smallest carrier unit and use
equal-length irrelevant replacement, conflict replacement, and position
permutations.  Deleting task facts, user preferences, or tool configuration
along with $K_1$ destroys identification.

\paragraph{The zero proves absence.}
It does not.  Different failure mechanisms converge on zero, and the strongest
artifact has a stateless ceiling.  The result is only that this protocol
observed no trial satisfying all five conditions.

\section{Interpretive and Governance Scope}
\label{app:scope}

The engineering distinction in this paper is between executing a fixed
determination and revising a determination after the system registers its
failure.  Philosophical discussions of judgment, rule following, explication,
and revisability motivate that distinction
\citep{kant1998,wittgenstein2009,carnap1950,quine1951,sellars1997,hegel2018},
but a benchmark score cannot resolve those debates.  Criterion revision is an
observable systems construct here, not a proxy for consciousness, autonomy,
understanding, or moral status.

Writable agent state also moves the practical boundary beyond frozen model
weights.  Once cross-episode state can change, attribution depends on what was
stored, who selected the write, whether the state persists, and whether a
matched intervention on that carrier changes later behavior.  Correct answers
alone cannot locate the boundary because a model may reconstruct the target
rule from the current input, as the \cmb{} calibration demonstrates.

Governance requirements remain distinct from the measured capability.  A
system specification should state who may write, which criterion layer may
change, which constraints remain immutable, and how revision, escalation,
audit, and rollback operate.  Local task criteria can be revisable while
permission and data boundaries remain fixed.  The protocol evaluates a narrow
revision mechanism; it does not authorize unbounded self-modification or imply
that every governing constraint should itself be writable.

\end{document}